\documentclass[runningheads]{llncs}
\usepackage{graphicx}
\begin{document}
\title{Optimisation of a Siamese Neural Network for Real-Time Energy Efficient Object Tracking}
%\title{Contribution Title\thanks{Supported by organization x.}}
%
\titlerunning{Optimisation of a Siamese Neural Network}
\author{Dominika Przewlocka\orcidID{0000-0002-5836-8604} \and 
Mateusz Wasala\orcidID{0000-0002-8631-8428} \and
Hubert Szolc\orcidID{0000-0003-3018-5731} \and 
Krzysztof Blachut\orcidID{0000-0002-1071-335X} \and 
Tomasz Kryjak\orcidID{0000-0001-6798-4444}} 

% \author{Dominika Przewlocka \and 
% Mateusz Wasala \and
% Hubert Szolc \and \\
% Krzysztof Blachut \and 
% Tomasz Kryjak} 
%
\authorrunning{D. Przewlocka et al.}
\institute{Embedded Vision Systems Group, Computer Vision Laboratory, \\ Department of Automatic Control and Robotics, \\ AGH University of Science and Technology, Krakow, Poland
\email{\{dprze,wasala,szolc,kblachut,tomasz.kryjak\}@agh.edu.pl}}
\maketitle              % typeset the header of the contribution
\begin{abstract}

% W~artykule przestawiono badania nad optymalizacją algorytmu śledzenia obiektów z wykorzystaniem syjamksiej sieci neuronowej dla potrzeb implementacji we wbudowanych systemach wizyjnych.
% Założono, że śledzenie ma być realizowane w czasie rzeczywistym, najlepiej dla strumienia wideo o wysokiej rozdzielczości, przy jak najmniejszym zużyciu energii. 
% Zastosowano takie techniki jak redukcja precyzji obliczeń oraz pruning.
% Wykorzystano narzędzie Brevitas, które jest przeznaczone do optymalizacji sieci dla potrzeb implementacji w układch FPGA. 
% Przetestowano szereg wariantów uczenia z różnym stopniem optymalizacji pamięciowo obliczeniowej. 
% Następnie przebadano jak użyte optymalizacje wpływają na skuteczność śledzenia dla testowanych sieci.
% Uzyskane wyniki pokazują, że stosując kwantyzację oraz pruning można znacznie zredukować złożoność pamięciową i obliczeniową sieci i tym samym użyć ją we wbudowanym systemie wizyjnym.
% TODO: TBC: dopisać końcówkę jak już będziemy mieć wyniki

In this paper the research on optimisation of visual object tracking using a Siamese neural network for embedded vision systems is presented.
It was assumed that the solution shall operate in real-time, preferably for a high resolution video stream, with the lowest possible energy consumption.
To meet these requirements, techniques such as the reduction of computational precision and pruning were considered. 
Brevitas, a~tool dedicated for optimisation and quantisation of neural networks for FPGA implementation, was used.
A~number of training scenarios were tested with varying levels of optimisations -- from integer uniform quantisation with 16 bits to ternary and binary networks.
Next, the influence of these optimisations on the tracking performance was evaluated. 
It was possible to reduce the size of the convolutional filters up to 10 times in relation to the original network.
The obtained results indicate that using quantisation can significantly reduce the memory and computational complexity of the proposed network while still enabling precise tracking, thus allow to use it in embedded vision systems. 
Moreover, quantisation of weights positively affects the network training by decreasing overfitting.

 %This work presents an analysis of several optimisation techniques to reduce memory and computational complexity of network, while maintaining a high accuracy of tracking. We use the Brevitas tool, which aims to optimise networks for FPGA implementation.

\end{abstract}

% --------------------------------------------------------------------------------------------------------------------
\section{Introduction}
Visual object tracking is one of the most complex tasks in computer vision applications and despite the~great number of algorithms developed over the last decades, there is still plenty of room for improvement. 
Well known challenges, typical for this task, involve object appearance changes (size, rotations), variable dynamics and potential similarity to the background, illumination changes, as well as object disappearing.
To handle the enlisted situations, advanced algorithms have to be used.
On the other hand, such systems are often used in edge devices like smart cameras used for video surveillance and vision systems for autonomous vehicles (cars, drones).
This rises another challenges: real-time processing and energy efficiency requirements, as well as support for high resolution video stream -- Full High Definition (FHD - $1920 \times 1080$) and Ultra High Definition (UHD -- $3840 \times 2160$). 

\begin{figure}
\includegraphics[width=\textwidth]{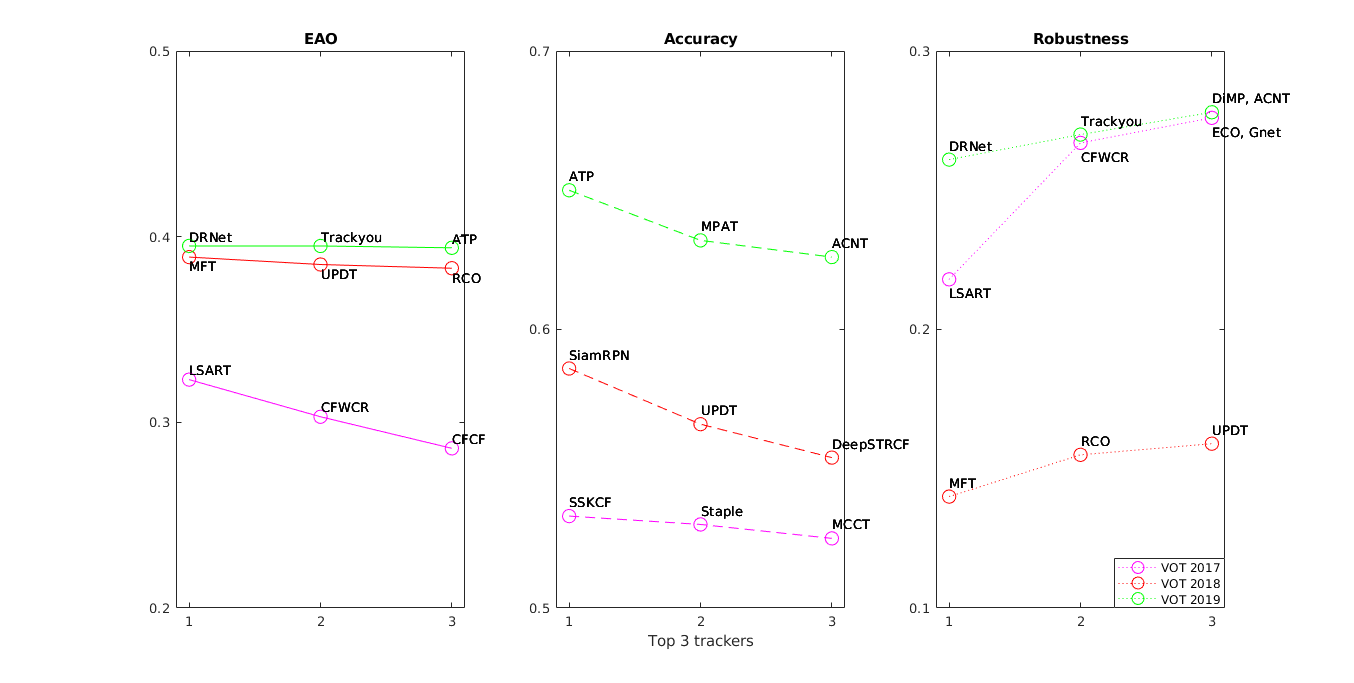}
\caption{EAO (Expected Average Overlap), Accuracy and Robustness of the best 3 trackers from three last editions of VOT: 2017, 2018 and 2019 in the baseline category.} \label{fig::history_vot}
\end{figure}

Following the subsequent editions of the well known in the community VOT (Visual Object Tracking) challenge, one can observe a~noticeable progress in accuracy of the proposed trackers over the last few years.
In the VOT challenge the trackers are evaluated using accuracy, robustness and the expected average overlap (EAO) which combines the two previous factors to reasonably compare different algorithms. 
Figure \ref{fig::history_vot} shows results of the top 3 trackers from the last three editions of the challenge in the baseline category. 
From 2017 to 2019 the EAO value of the best tracker improved by 18.23\%, which has to be considered as a~significant progress.

Many recently proposed solutions either do not operate in real-time or are accelerated on rather energy inefficient GPUs. 
Their accuracies places them high in VOT (Visual Object Tracking) challenges, thus makes them appropriate for applications with a~great need for precise tracking, such as autonomous vehicles (AV), advanced driver assistance systems (ADAS) or advanced video surveillance systems (AVSS). 
There is therefore a~necessity for using them on edge or embedded devices with limited power budget. 

% pc, gpu, fpga
However, the computational complexity of the state-of-the-art trackers is so demanding, that in order to achieve real-time processing, high-end GPUs are needed.
For many applications this is often unacceptable, since GPUs consume a~significant amount of energy: for example autonomous vehicles powered by accumulators need low-power solutions working in real-time. 
There is therefore a~need for lightweight algorithms, in terms of memory and computational complexity, that are possible to accelerate on embedded devices. B
It is obvious that such trackers still need to provide the best possible accuracy, so research in this area has to focus on optimisation of the state-of-the-art solutions.
One of the possible choices for the target computing platform are FPGAs (Field Programmable Gate Arrays) or heterogeneous devices (combining programmable logic and CPUs), as they enable high level of parallelisation, while consuming relatively low energy (a~few watts). 

% krótko o syjamskich - obrazek poglądowy: nazewnictwo exemplar-instance
Lately, many of the best trackers use Siamese Neural Networks. 
A~Siamese network is a~Y-shaped network with two branches joined to produce a~single output. 
The idea itself is not new, as it originated in 1993 in works about fingerprint recognition \cite{siamfinger1993} and signature verification \cite{siamsign1993}. 
A~Siamese network measures the similarity of two processed inputs, thus it can be considered as a~similarity function. 
In fact, many of the Siamese-based trackers rely solely on that assumption. 
The exemplar image of an object (from the first frame) is treated as the first input to the network, while the region of interest (ROI), where the object might be present in the consecutive frame, forms the second one. 
The network searches for similar to the object's appearance areas in the search ROI. 
Depending on specific tracker, the output is a~new bounding box or just the location of the object (the centre of mass) that has to be further processed to obtain the bounding box.

% main contribution TODO
The presented research focuses on adapting a~state-of-the-art tracker based on Siamese Neural Networks to embedded devices. 
We aim to optimise the network for memory and computational complexity which shall directly translate into meeting real-time requirements via an FPGA implementation. 
The main contribution of this paper is the research on optimising Siamese Neural Networks for object tracking, by reducing the computational and memory complexity.
To the authors' best knowledge no prior work described the results of such various quantisations -- from uniform integer quantisation of 16, 4 bits, ternary and binary filters. 

The rest of the paper is organised as follows: Section \ref{sec::previous} briefly describes previous work in related domains -- deep learning, tracking and acceleration of computations. 
Section \ref{sec::tracking} describes tracking using Siamese Neural Networks. 
In Sections \ref{sec::Brevitas} and \ref{sec::optimisations} tools and optimisation techniques for reducing the computational and memory complexity of neural networks are presented. 
Ultimately, Section \ref{sec::experiments} describes the conducted experiments and Section \ref{sec::results} the obtained results. 
Finally, in Section \ref{sec::summary} the presented research is summarised and possible future work is discussed.

% --------------------------------------------------------------------------------------------------------------------
\section{Previous work} 
\label{sec::previous}

Recently several trackers based on Siamese Neural Networks were proposed by the scientific community.
In \cite{goturn2016} the authors proposed a~tracker based on a~Siamese Neural Network named GOTURN.
The two branches are based on the Alexnet \cite{alexnet2017} DCNN and the network is trained for bounding box regression. Given two input images, one as an exemplar of object to track and the second with the ROI, the network predicts the location of the object in the new frame. 
This algorithm was implemented using Nvidia GeForce GTX Titan X GPU with cuDNN acceleration and achieved the speed of 6.05 ms per frame (165 fps) and GTX 680 GPU with the speed of 9.98 ms (100 fps). 
The tracker is able to process 2.7 frames per second on a~CPU. 

In \cite{siamfc2016} the authors proposed a~slightly different solution. 
Based on the two inputs (exemplar and instance -- ROI), the Siamese Network estimates the location of object in the ROI. 
The possible variance of the object's size is handled by analysing multiple scales, thus calculating the network's output multiple times for a~single frame. 
Tracking implemented on computer with Nvidia GeForce GTX Titan X and an Intel Core i7-4790K at 4.0GHz runs at 86 and 58 fps with respectively 3 and 5 scales.

The direct continuation of the work \cite{siamfc2016} resulted in \cite{siamcf2017}. 
The authors presented a~tracker which combines a~Siamese Neural Network and correlation filters.
% The correlation filter was interpreted as a~differentiable layer in the neural network. 
The results showed that it allowed to achieve state-of-the-art performance with lightweight architectures.
Using the same hardware, the tracker achieves a~speed of 75-83 fps (depending on network's depth). 

In \cite{dsiam2017} the authors presented a~Dynamic Siamese Network that allows for online learning of the target appearance from previous frames. 
The network's output is a~score map (similar as in \cite{siamfc2016}) that indicates the centre of the object's location. The tracker achieves 45 fps on a~Nvidia Titan X GPU.

Another interesting idea was presented in \cite{siamban2020}, where the network is trained for bounding box regression but without any prior knowledge abut the bounding box (there are no pre-defined anchor boxes nor multi-scale searching). 
This solution achieved a~speed of 40 fps on Nvidia GTX 1080Ti and Intel Xeon(R) 4108 1.8GHz CPU with AUC equals to 0.594 on OTB100 benchmark.

A~comparison of the above-mentioned Siamese-based trackers evaluated on OTB 2013 benchmark is presented in Table \ref{tab::trackers}. 
Moreover, based on the results from the VOT 2019 challenge \cite{vot2019} 21 (37\%) of participating trackers were based on Siamese Networks, from which 3 made it to the top ten best trackers according to the EAO measure. 
% TODO: VOT19: jak znaleźć artykuły o tych trzech algorytmach? SiamMargin A.43, SiamFCOT A.42, SiamCRF A.37

\begin{table}[!t]
\caption{Comparison of trackers based on Siamese Neural Networks. The choice of the benchmark OTB 2013 was dictated by reports in articles provided by the authors.}\label{tab::trackers}
\centering
\begin{tabular}{c c c c }
\hline
Tracker & AUC in OTB 2013 { } & { } Speed (fps) { } & { } Acceleration \\
\hline
GOTURN \cite{goturn2016} & 0.447 & 165 & Nvidia Titan X GPU\\
SiamFC-3s \cite{siamfc2016} & 0.607 & 86 & Nvidia Titan X GPU\\
DSiam \cite{dsiam2017} & 0.642 & 45 & Nvidia Titan X GPU\\
CFNet-conv2 \cite{siamcf2017} & 0.611 & 75 & Nvidia Titan X GPU\\
\hline
\end{tabular}
\end{table}

% podsumować to GPU
All the above mentioned algorithms have high tracking accuracy and operate in real-time with the use of high-end GPUs. 
Hence, if one wishes to use them in low energy systems, implementation on embedded devices is necessary. 
One of the possible solutions is to use FPGA devices, as they provide the possibility of computation parallelisation and are energy efficient. 
The currently proposed solutions have high memory and computational complexity and usually it is impossible to fit such networks entirely on the considered device. 
Regardless of tracking, several simplifications for convolutional neural networks were proposed: binary or XNOR networks with zero-one weights and activations \cite{xnor2016}, ternary (\{-1, 0, 1\}) networks \cite{gxnor2017}, networks with other quantisations (uniform integer quantisation) adopted a~priori \cite{quant2016}, as well as trained \cite{dcompr2016}, or pruning, i.e. reducing networks' parameters by removing neurons or connections between them \cite{prun2020}. 
Moreover, recently there is a~strong trend for research on efficient network design \cite{shufflenet2018}. 

There are few papers covering acceleration of Siamese Neural Networks for tracking.
In \cite{tinysiam2019} the authors presented a~tiny Siamese Network for visual object tracking, that due to the low computational complexity can be implemented in embedded devices. 
The presented algorithm operates with the speed of 16 fps on a~CPU and 129 fps on a~GPU and is fairly comparable to other state-of-the-art solutions based on Siamese Neural Networks (like SiamFC or GOTURN \cite{siamfc2016,goturn2016}) by achieving the score of AO (average overlap) equal to 0.287 (for example, SiamFC achieves 0.349 AO with 39 fps on a~GPU).

Finally in \cite{minitracker2018} the authors presented an FPGA implementation of lightweight system for visual object tracking using a~Siamese Neural Network. 
Basing on the \cite{siamfc2016} solution, the authors developed a~network with a~more regular architecture (with filters of the same kernel sizes) and then performed pruning and quantisations. 
This allowed to achieve results comparable, in context of precision or AUC, to the original solution.
The tracker operates with the speed of 18.6 fps on a~ZedBoard and is highly energy efficient with only 1.284W.
%TODO tu albo później jakoś do tego trzeba się odnieść :)

Despite the great interest in the area of tracking based on Siamese Neural Networks, not much research is concentrated on accelerating this excellent trackers in order to use them in systems with low power requirements. 
Therefore, we believe that the results of our experiments fill a~gap in this domain.

%there is a gap to fill with extensive research and present the  in that domain.

% --------------------------------------------------------------------------------------------------------------------
\section{Tracking with Siamese Neural Networks} \label{sec::tracking}
% krótko o syjamskich - obrazek poglądowy: nazewnictwo exemplar-instance
% opis algorytmu, na którym bazujemy

\subsection{Siamese Neural Networks}

A~Siamese network is a Y-shaped network with two branches joined to produce a~single output. % -- exemplar architectures are shown in Fig. \ref{fig::siamese_types}. 
Irrespective of the branches' structure, a~Siamese network can be considered as a~similarity function that measures the resemblance between two inputs.  
% \begin{figure}[!t]
% \includegraphics[width=\textwidth]{figures/DP_types_of_siamese_nets.png}
% \caption{Different types of Siamese networks based on \cite{pflugfelder2017}: (a) Two-channel, (b) Pseudo Siamese with shared weights across branches, (c) Siamese, (d) Two-Stream, (e) Recurrent} \label{fig::siamese_types}
% \end{figure}
We can define the basic Siamese architecture using Equation (\ref{eq::siamese}), where $\phi$ stands for extracting deep features (e.g. via network's branches), $z$ represents the template (e.g. exemplar, object to track), $x$ represents the patch that is compared to the exemplar (e.g. search region, ROI) and $\gamma$ is a~cross-correlation.
\begin{equation} \label{eq::siamese}
y = \gamma (\phi(z), \phi(x))
\end{equation}

\subsection{Tracking}

\begin{figure} [!t]
\centering
\includegraphics[width=0.7\textwidth]{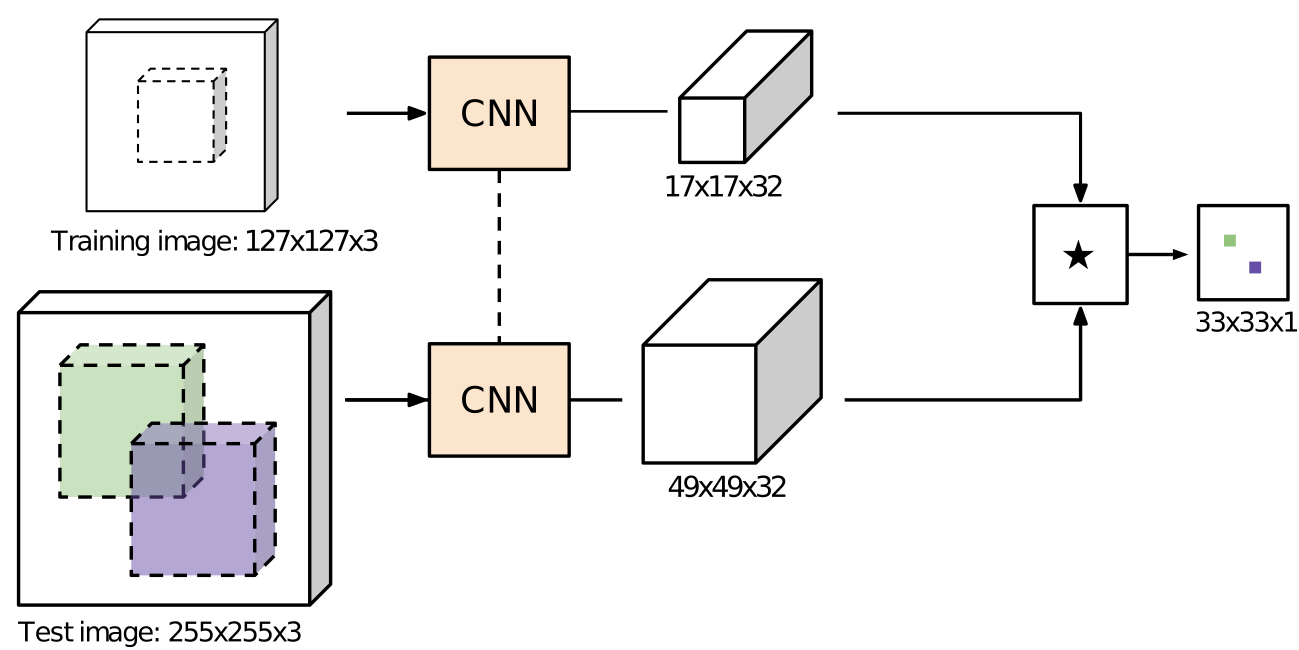}
\caption{A fully-convolutional Siamese Network for tracking \cite{siamfc2016}.} \label{fig::siamfc}
\end{figure}

As discussed in Section \ref{sec::previous}, several trackers based on Siamese Networks exist. 
In this research we focus on the solution presented firstly in \cite{siamfc2016} and then modified in \cite{siamcf2017}, mainly due to its good performance and simplicity (which is important in context of FPGA implementation).
Its scheme is presented in Fig. \ref{fig::siamfc}.
The input $z$ represents the tracked object selected in the first frame and scaled to 127x127 pixels.
In the next frames, the ROI around the previous object's location is selected and presented to the network as the input $x$, scaled to 255x255 pixels. 
Both patches (exemplar and ROI) result in two blocks of feature maps, of sizes 17x17x32 (for $z$) and 49x49x32 (for $x$) respectively.
Finally they are cross-correlated and a~heat map of size 33x33 is obtained. 
Its highest peak indicates the object's location.  
The ROI is selected approximately 4 times greater than object's size, over five scales to handle object's size variations. 

% --------------------------------------------------------------------------------------------------------------------
%TODO Dodać cytowanie FINN
\section{The used tools: Brevitas and FINN} 
\label{sec::Brevitas}
% co to jest i po co
In order to implement deep learning algorithms in FPGA devices it is highly advisable to somehow optimise the network's architecture. 
This increases the performance and, in some cases, even allows to only use internal memory resources.
Apart from designing lightweight networks, there are several other options to compress it's parameters, like quantisation and pruning. 
However, these methods are not independent of the training process. 
For a~long time most popular tools for deep learning, such as Tensorflow or Pytorch standalone, did not contain such extensions and no open source tool was able to perform such optimisations. 
However, as the acceleration of deep networks on embedded devices gained on popularity in last few years, this situation changed. 
This resulted in several interesting tools for deploying them on FPGA/SoC platforms (Field Programmable Gate Array/System on Chip), e.g. Caffe to Zynq from Xilinx reVISION stack, Vitis AI or open source Brevitas/FINN. 
In this work we will focus on the last solution. 
% \begin{figure}
% % TODO: może zrobić analogiczną ale w poziomie? strasznie dużo miejsca zajmuje...
% %TODO TK: można z tym poczekąć.
% \centering
% \includegraphics[width=0.4\textwidth]{figures/DP_finn_stack.png}
% \caption{Brevitas and FINN framework for DNN acceleration \cite{finn2020}.} \label{fig::finnstack}
% \end{figure}

% Brevitas and FINN are complementary tools (see Figure \ref{fig::finnstack}), but as they are still under development, not all features are yet ready. 
Brevitas and FINN are complementary tools, but as they are still under development and not all features are yet ready. 
Brevitas is a~Pytorch library that enables the training of quantised networks, so it allows optimising the considered algorithm. 
The trained model can be then compiled into a~hardware design using the FINN tool. 
However the Brevitas-to-FINN part of the flow is currently available for a~very constrained set of network layers. 
The current version of the framework is sufficient for designing algorithms ready for hardware implementation in the future releases of FINN or ``manually''.

\section{Optimisation Methods} \label{sec::optimisations}
% wymienić wszystkie metody brevitasa i nie tylko (pruning)

\subsection{Quantisation}

Quantisation allows to reduce the number of bits needed to store values, which can be (and often is) critical for deploying networks on embedded devices. 
Both weights and activations can be quantised. 
Despite several approaches involving quantising weights after training, the best results are obtained when training the network with reduced precision parameters. 
%Moreover, the optimal bit-width can also be learned during the training process. %TK Na razie tego nie używamy.
Using Brevitas, one can set both type of quantisation (uniform integer quantisation, binary or ternary), as well as particular bit width for weights and activations.

\subsection{Pruning}
The idea of pruning is as old as 1990, when Yan LeCun \cite{lecun1990} showed that removing unimportant weights from a~neural network can result in better generalisation, fewer training examples needed and acceleration of training, as well as inference. 
A~neural network is a~rather redundant structure and by finding the neurons that do not contribute much to the output, its architecture can be optimised. 
One possible way to do that is to rank the neurons and search for those which are less relevant -- according to L1/L2 norm of weights in a~neuron, mean activations and other measures. 
In practice, the process of pruning is iterative with multiple fine-tuning steps: after suppressing low ranked neurons, the network's accuracy drops and a~retraining step is necessary. 
Usually the network is pruned several times, deleting only a~relatively small number of neurons at once. 

% -------------------------------------------------------------------------------------------------------------------
\section{Experiments} \label{sec::experiments}

The proposed network is similar to the one presented in \cite{siamfc2016}. 
In order to optimise it, several quantisation experiments were performed. 
It was decided to focus more on the hidden layers than on the input and output ones, as while conducting preliminary tests it turned out that radical quantisations of all layers causes difficulties in training. 
Using weights with higher precision for the last layers results in better accuracy of tracking, which is showed in experiments.
Moreover, based on similar observations, the activation layers also operate with higher precision (with uniform integer quantisation).

\subsection{Training}
% w jaki sposób zaimplementowano uczenie
% pytanie czy czas use, czy have used ? 
In order to obtain representative and comparable to each other results, we have defined the same training parameters for all test cases. 
The networks were trained using the ILSVRC15 dataset. 
For the presented experiments, approximately 8\% (that is 10000 image pairs of object and ROI) of the dataset was used for training. 
The authors of original Siamese Network compared the accuracy -- measured as average IOU -- of their tracker while training on different portions of the ImageNet dataset. 
In that case the 8\% evaluation resulted in an 48.4\% accuracy (for the whole dataset: 52.4\%).
Thus, to reduce the training time, we have followed the same approach -- in our case one training experiment took about 12 hours on a~PC with Nvidia GeForce GTX 1070 GPU.
The number of epochs was set to 30.
Naturally, when training the network one should adjust its architecture to the task and available data to prevent the network from learning ''by heart''.
Nevertheless, in this case, when deliberately training on a small subset of the dataset, this could happen, thus number of epochs was minimised to the value that allowed for a~decent evaluation while preventing from overfitting. 
After these preliminary experiments, the whole process can be repeated for a~one chosen network (with the best relation of computational and memory complexity to accuracy) on the whole dataset to obtain better results. 

\begin{table}[!t]
\caption{Quantisations of the proposed Siamese Neural Network for tracking. FP stands for floating point precision, INT for uniform integer quantisation, TERNARY and BINARY respectively for two and one bits precision.}\label{tab::quantizations}
\centering
\begin{tabular}{ c c c c c c c c c c c }
\hline
No{ } & { }First Layer{ } & { }Hidden Layers{ } & { }Last Layer{ } & { }Activation{ } & { }Convolutional Parameters{ } \\
\hline
1 & FP 32 & FP 32 & FP 32 & FP 32 & 85.5 MB \\
% 2 & INT 32 & INT 32 & INT 32 & INT 32 & 85.5 MB \\
2 & INT 16 & INT 16 & INT 16 & INT 32 & 38.3 MB \\
3 & INT 16 & INT 4 & INT 8 & INT 32 & 13.4 MB \\
4 & INT 16 & TERNARY & INT 8 & INT 32 & 9.3 MB \\
5 & INT 16 & BINARY & INT 8 & INT 32 & 7.2 MB \\
\hline
\end{tabular}
\end{table}

The conducted experiments are summarised in Table \ref{tab::quantizations}.
They were selected to evaluate how the reduced precision affects the training progress and it's results.
The first case represents the network with floating point precision and is treated as the baseline. 
The last column shows the memory needed for convolutional filters (other parameters are constant for all test cases). 
It is worth noting that on FPGA devices the use floating point computations is generally more complex, resource consuming and thus less energy efficient than fixed-point ones.

% ------------------------------------------------------------------------------------------------
\section{Results} \label{sec::results}
% tabelki i porównania
% czas inferencji - komentarz ?
% skuteczność

The considered system was evaluated twofold -- during training of the network and object tracking.
In the first case two metrics are considered -- the value of the loss function and the average centre error (the difference in pixels between the object's location returned by the network and the ground truth).
\begin{figure}[!t]
\centering
\includegraphics[width=0.9\textwidth]{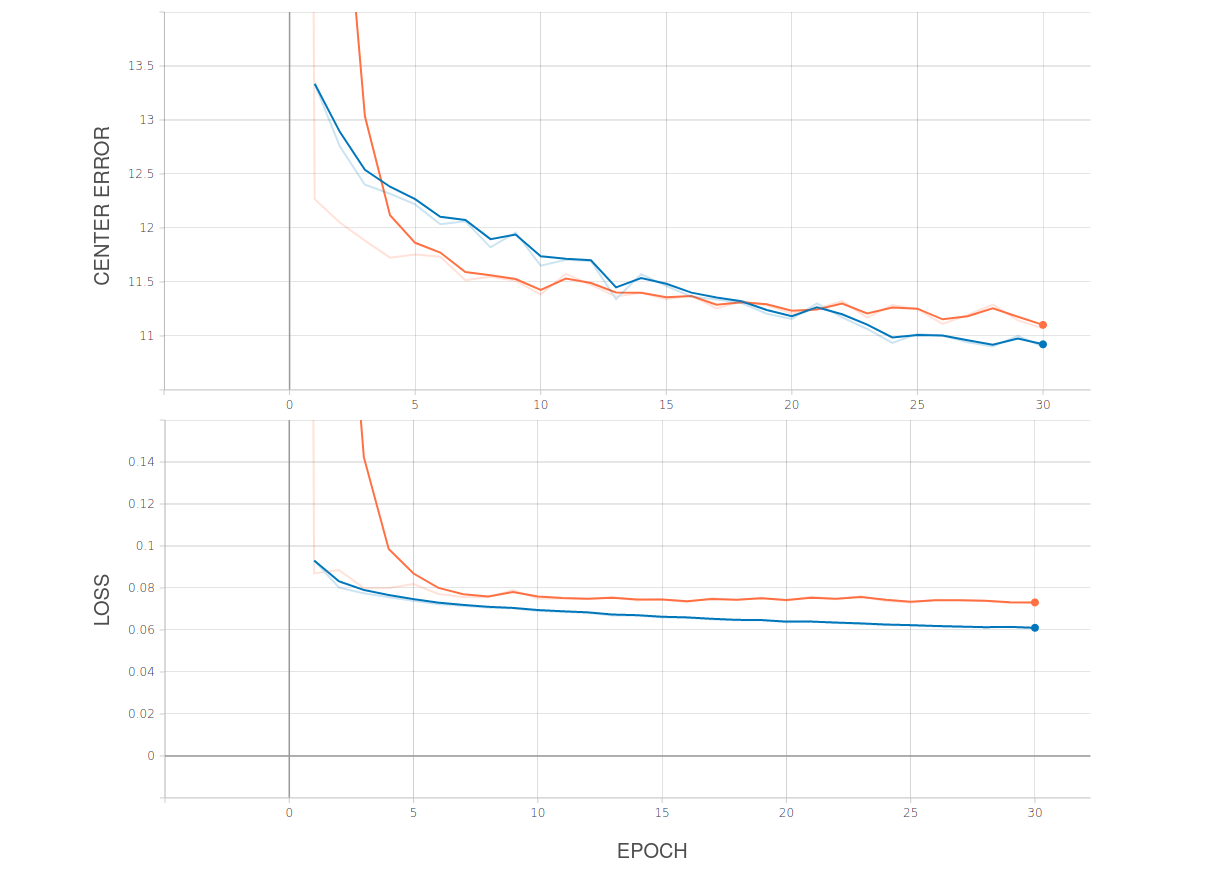}
\caption{The training progress for the baseline network with floating point precision after 30 epochs. The upper plot shows the centre error, while the lower the loss function. The blue line represents training, while the orange one validation.} \label{fig::training_progress}
\end{figure}

\begin{figure}[!t]
\centering
\includegraphics[width=0.5\textwidth]{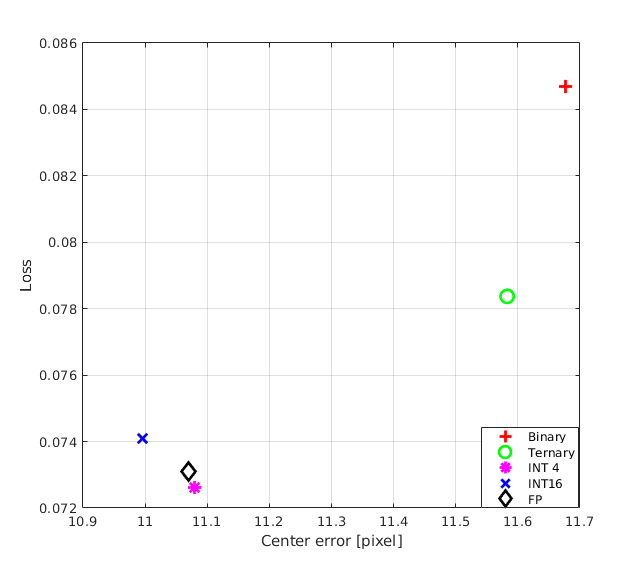} 
\caption{Comparison of training results for the best evaluation epoch for different quantisations. The differences are small because of the low number of epochs, however the trend is maintained.} \label{fig::training_comparison} %KB comparison of training results for best... (bo to nie jest porównanie treningu, tylko jego wyników)
\end{figure}

In Figure \ref{fig::training_progress} an exemplar training progress for the baseline floating point network is shown. 
Due to the small differences between the tested networks, the training results are presented jointly in the Fig. \ref{fig::training_comparison} instead of presenting each training curve separately. 
As expected, the networks with lower precision have greater centre error and loss value (thus lay in upper right corner), on contrary to those with higher precision. 
The INT16 network slightly outperforms other ones, having a~lower centre error. 
This situation is caused by a different quantisation in the last layer. %KB na pewno INT16? ten błąd jest nieznacznie mniejszy. i chyba outlier to tylko rzeczownik, więc tu nie może być czasownikiem?
%TK Zgoda z KB zamniłem na slighlty outperforms. Natomiast wytłumaczenie jest dla mnie niejasne ?

The tracking results are presented in Table \ref{tab::tracking_accuracy}. 
The tracker was evaluated on a~dataset constructed from TempleColor, VOT14, VOT16 set (without OTB sequences -- same as in the original paper).  
The tracking precision is calculated as the ratio of the number of frames with centre error below the predetermined threshold (in our case equal to 20) in relation to the length of the given sequence. 
The analysis of the average precision values allows to draw interesting conclusions, while comparing the quantised versions to baseline floating point one. 
%First of all, there is a~dramatic drop of precision between the baseline and the INT32 implementation. 
First of all, there is a~precision growth with incerasing quantisation (excluding the INT16 network) with binary network achieving the precision of 44.22 on the top.
A~similar tendency can be observed using the IOU (Intersection Over Union) metric. 
The obtained results are not as good as in the original paper, nevertheless this situation should be solved while training network for a~longer time (as planned in the future work).
It is essential to notice that the binary tracker is relatively better than the one using floating point computations.
However, if we acknowledge that neural networks are generally redundant structures, this behaviour is not so astounding, as by quantising weights, the overfitting is prevented. 
A~similar phenomenon was observed in \cite{minitracker2018} or \cite{han2015}.

\begin{table}[!t]
\caption{Average tracking precision and average IOU for trackers using quantised Siamese Networks.}\label{tab::tracking_accuracy}
\centering
\begin{tabular}{ l c c }
\hline
Network{ } & { }Precision [\%]{ } & { }IOU [\%]{ }\\
\hline
 FP & 38.58 &  28.42\\
%  INT32 & 19.27 & 14.76 \\
 INT16  \footnote{1} & 42.87 & 31.24 \\
 INT4 & 36.73 & 27.83 \\
 TERNARY & 44.35 & 33.40 \\
 BINARY & 44.22 & 32.59\\
\hline
\end{tabular} \\
\end{table}

\footnotetext[1]{Quantisation of the last layer for network INT16 is lower than in other scenarios (INT16 vs INT8).}

One experiment was performed with lower output layer quantisation -- the network INT16 is using uniform integer quantisation with 16 bits for all convolutional layers. 
The results show that the precision reduction in the last layer has greater influence on network.

\subsubsection{Inference profits}

Regardless the lower memory requirements, such quantisations could have impact on the inference time. 
Given ternary or binary networks, the computational complexity is reduced via replacing standard MAC (Multiply and Accumulate) operations with their binary equivalents. 
Such operations, if properly implemented, lead to lower execution time. 
In the case of FPGA implementation it also results in lower resources usage.

% -------------------------------------------------------------------------------------------------------------------
\section{Summary} \label{sec::summary}
% podsumowanie wyników
% koniecznie odniesienie, że docelowo to ma działać r-t dla 4K
% future work: włożenie sieci do całego algorytmu śledzenia i sprawdzenie jak wpływa to na działanie

The presented research explores the possibilities of reducing the computational and memory complexity of a~Siamese Neural Network tracker. 
Two experiments were performed -- one during network training and evaluation and the second during tracking.
In the first case the quantisation had a~negative, as expected, impact on the loss function and centre error.
It also turned out that it allowed to reduce the network's tendency to overfit. 
In the second case, reducing the precision solely in the hidden and slightly in other layers, that is minimising the size of the network in MB, resulted in the increase of the tracker performance.
This trend should be further investigated and confirmed on a bigger dataset.
The later property is extremely important when aiming for FPGA implementation and enabling highly accurate real-time tracking for a~UHD video stream, as a~binary network requires more than 10 times less memory than a~floating point one.

%The obtained results show that for the used setup, quantising the Siamese neural networks allows to increase the efficiency of the tracking algorithm. 

%However, the influence of quantisations on the network's training shall be further investigated as it seems 

%After firmly  
%However, the obtained results have to be repeated with bigger dataset to confirm the trend.

%TODO Tu by przydało się zadanie konkrentu z jakimiś %.

For future work we plan to perform additional tests using greater number of training data for networks with ternary and binary weights, as the obtained results reveal their tracking potential. 
For these networks the pruning experiments shall also be executed, as they could allow for even better optimisation. 
Moreover, more careful analysis on quantisation of input and output, as well as activation layers is needed since such reduction would enable even less memory complexity. 
Eventually, the final version of the tracker has to be tested on the sequences from the newest VOT challenges for better comparison with state-of-the-art approaches.
Finally, the solution should be implemented on a~FPGA device.

\subsubsection*{Acknowledgements} 
% Removed for blind review.
%TODO Ew. Grant Dziekański DP ?
% BLIND REVIEW
The work presented in this paper was supported by the Faculty of Electrical Engineering, Automatics, Computer Science and Biomedical Engineering Dean grant -- projet number XXXX (first author) and the National Science Centre project no. 2016/23/D/ST6/01389 entitled ''The development of computing resources organization in latest generation of heterogeneous reconfigurable devices enabling real-time processing of UHD/4K video stream''.

\bibliographystyle{splncs04}
%\bibliography{mybibliography}

\end{document}